\documentclass[conference]{IEEEtran}
\IEEEoverridecommandlockouts
\usepackage{cite}
\usepackage{amsmath,amssymb,amsfonts}
\usepackage{algorithm,algpseudocode}
\usepackage{graphicx}
\usepackage{multirow}
\usepackage{textcomp}
\usepackage{xcolor}
\usepackage{dsfont}
\usepackage{url}
\def\BibTeX{{\rm B\kern-.05em{\sc i\kern-.025em b}\kern-.08em
    T\kern-.1667em\lower.7ex\hbox{E}\kern-.125emX}}
\begin{document}
\bstctlcite{IEEEexample:BSTcontrol}

\title{Federated Deep Clustering Networks for High-Dimensional and Heterogeneous Data

\author{\IEEEauthorblockN{Morris Stallmann, Charalampos S. Kouzinopoulos, Marcin Pietrasik, Anna Wilbik}
\IEEEauthorblockA{\textit{Department of Advanced Computing Sciences (DACS)} \\
\textit{Maastricht University}\\
Maastricht, Netherlands \\
\{m.stallmann, charis.kouzinopoulos, marcin.pietrasik, anna.wilbik\}@maastrichtuniversity.nl}}
%\title{Federated Deep Clustering Networks: Clustering High-Dimensional and Heterogeneous Data in Federated Learning \\

%\thanks{Identify applicable funding agency here. If none, delete this.}
}

\maketitle

\begin{abstract}
Clustering high-dimensional data is a fundamental task in unsupervised machine learning with applications to a variety of domains. In the centralized data scenario, this task is commonly solved using deep clustering methods that utilize deep neural network architectures to learn clustering-friendly latent space representations.
In Federated Learning, where data is distributed between clients and is private, deep clustering methods are less explored. In particular, recently introduced federated deep clustering methods, despite showing very promising performance, still fall short in reliably providing good performance if data across clients are non-identically-independently distributed.
In this work, we introduce a generalization of Deep Clustering Networks to the federated scenario, named FedDCN, that simultaneously optimizes a reconstruction loss and a clustering loss. To ensure robustness and latent space alignment in non-identically-independently distributed data scenarios, FedDCN generates synthetic data augmentations, and its learning objective includes a geometric regularization for latent space alignment. Through experimental evaluation, the effectiveness of the approach under IID and non-IID assumptions is demonstrated, and future research directions are identified.
\end{abstract}


\section{Introduction}
Clustering is the process of partitioning objects in a data space such that objects in the same partition are similar to each other and different from objects in other partitions. It helps to organize data in the absence of labels and has been applied to various use cases, such as fraud detection \cite{de_roux_tax_2018}, image segmentation \cite{mittal_comprehensive_2022}, acoustic scene identification \cite{li_acoustic_2020}, video understanding \cite{peng_unsupervised_2020}, as well as to a number of application areas including logistics, manufacturing, energy, and healthcare \cite{oyewole_data_2023}.
In many real-world deployments, data are inherently decentralized and remain distributed across multiple clients. Aggregating such data at a central repository is often impractical due to communication and storage constraints as well as privacy considerations.

Federated Learning (FL) addresses these challenges by collaborative model training over decentralized datasets while keeping raw data local to each client. Participating clients iteratively communicate model updates, thus mitigating privacy risks and other constraints associated with data centralization.

Consequently, it is not surprising that federated clustering has attracted increasing attention and a number of methods have recently been proposed and extended to the federated setting (see Section \ref{sec:related_work}). Classic approaches such as federated versions of $k$-means or fuzzy $c$-means have been successfully generalized to the federated setting and show robust behavior even under the assumption of non-IID data. However, these ``classical" methods struggle with high-dimensional data \cite{winkler_fuzzy_2011, steinbach_challenges_2004}. In the non-federated setting, deep clustering methods are widely recognized as state-of-the-art when clustering high-dimensional data \cite{wei_overview_2024, zhou_comprehensive_2025}.
Deep clustering methods combine deep neural network (DNN) architectures for low-dimensional and clustering-friendly latent representation learning with grouping mechanisms to identify meaningful data partitions. 
To enable clustering of high-dimensional data in the FL setting, a number of recent works introduce federated deep clustering protocols. Despite very promising results, many of these pioneering approaches fail to reliably produce high-quality data partitions under the assumption of data heterogeneity across clients (see Section \ref{sec:rel_fed_deep_clustering}). Statistical data heterogeneity, or non-IIDness, refers to situations where the local datasets follow different distributions, and remains one of the key challenges in FL (see Section \ref{sec:rel_fed_non_iid}). In this work, we specifically study the impact of label skew and refer to it as non-IIDness throughout the paper.

This work aims to address one of the main challenges in federated deep clustering and move towards greater robustness against client data heterogeneity. To this end, we propose a federated protocol for learning Deep Clustering Networks (DCN) \cite{yang_towards_2017}, a well-known deep clustering architecture and optimization scheme in the centralized case that was introduced as an improvement over Deep Embedded Clustering (DEC) networks. DEC is the base for state-of-the-art federated deep clustering methods that this work aims to improve through the introduction of Federated Deep Clustering Networks (FedDCN).
To instill non-IID robustness, we propose a batch augmentation technique in combination with a geometric regularizer that complements the network's autoencoder architecture and is well grounded in existing research on aligning local and global distributions. 

%Moreover, since the centralized approach relies on a strong pretraining phase, we propose a method that reduces pretraining data requirements by introducing a geometry-regularized loss function. Based on our experimental evaluation, we draw the conclusion that .... 

The main contribution of this paper is a federated deep clustering framework that:
\begin{enumerate}
    \item is the first to generalize DCN to the federated setting.
    \item does not assume access to a centralized dataset for pretraining, unlike other approaches, thereby being more widely applicable in privacy-sensitive contexts.
    \item achieves state-of-the-art performance on benchmark datasets and robustness against data heterogeneity through synthetic data augmentation and a geometry-aware loss function.
\end{enumerate}

Section~\ref{sec:related_work} introduces related works; our proposed method is described in Section \ref{sec:proposed_method}; its experimental evaluation is described and discussed in Section \ref{sec:exp_eval} before concluding with final remarks in Section \ref{sec:concluding_remarks}.

\section{Related Work}\label{sec:related_work}
This section provides an overview of deep clustering and prior work on federated deep clustering, and discusses strategies for handling data heterogeneity in FL to clearly position our work within the existing literature.

\subsection{Non-Federated Deep Clustering}
%Deep clustering methods address a fundamental short-coming of traditional partitioning based (like $kt$-means), density-based (like DBSCAN) or hierarchy-based clustering methods (like agglomerative clustering): They do not work well in high-dimensional spaces, where distance becomes less meaningful and data tends to be sparse. 
Deep clustering methods utilize the representation learning capabilities of DNNs and learn clustering-friendly (low-dimensional) latent spaces from high-dimensional data spaces to address the shortcomings of traditional clustering methods.

A  taxonomy based on network architecture and optimization scheme is suggested in \cite{wei_overview_2024} to define four broad groups of deep clustering methods. %According to \cite{zhou_comprehensive_2025}, the methods can be grouped based on how they learn suitable representations and how they perform the partitioning of the input space.
The methods in the first group use DNNs to directly output latent representations and cluster assignments (DNN-based clustering).
The second group comprises methods that utilize autoencoders (AE) to learn a latent representation (AE-based clustering), and the clustering itself does not necessarily have to be performed by a DNN. Our proposed method falls into this group.
One of the fundamental methods in this category is DEC, introduced in \cite{xie_unsupervised_2016}. It is one of the first algorithms to explicitly optimize latent space representation mapping so that a clustering loss is minimized using stochastic gradient descent. 
%Specifically, the Kullback-Leibler (KL) divergence between (soft) cluster assignments and an auxiliary distribution (calculated using assignments and cluster frequencies) is optimized.
Later, DCN was introduced in \cite{yang_towards_2017} as an improvement of DEC. DCNs train AEs and a $k$-means model to simultaneously minimize reconstruction loss and $k$-means clustering loss. %To the best of our knowledge, our work is the first federated DCN generalization.
Methods that fall into the third group utilize Generative Adversarial Networks (GANs), where the adversarial training is augmented with a clustering loss. In CatGAN \cite{springenberg_unsupervised_2016}, for example, the discriminator has to decide whether a sample is real or fake and has to estimate its cluster assignment. 
Lastly, the fourth group consists of Graph Neural Network (GNN)-based methods that use GNNs to map graph nodes to a latent space where the clustering is performed subsequently. 
For an exhaustive overview, refer to recent surveys \cite{zhou_comprehensive_2025, wei_overview_2024}.

AE-based methods have an advantage over DNN-based methods in the federated setting: the encoder-decoder architecture can be utilized to address non-IIDness through the generation of synthetic data, as described in Section \ref{sec:proposed_method} below. GAN-based methods may have the same advantage, but are generally considered to be harder to train \cite{wei_overview_2024}. We did not consider GNN-based methods, since the input of this work is not graphs. 

\subsection{Federated Deep Clustering}\label{sec:rel_fed_deep_clustering}
%To study the state-of-the-art of federated deep clustering, we perform a small scale literature review.
Recently, a number of federated deep clustering methods have been introduced, of which the majority being DNN- or AE-based. 
One of the first AE-based federated deep clustering methods is a federated version of DEC (F-DEC) \cite{mashhadi_deep_2021}. F-DEC uses convolutional autoencoders in combination with a clustering layer and optimizes a weighted sum of Kullback-Leibler (KL) divergence (as clustering loss) and reconstruction loss. The approach assumes the availability of a centralized dataset for pretraining and achieves strong performance (accuracy and normalized mutual information, see Equation (\ref{eq:nmi})) on IID datasets. However, the performance decreases by up to 50\% on the MNIST dataset as non-IIDness increases. 
Similarly to F-DEC, FDEC, as proposed in \cite{xu_federated_2025}, minimizes KL divergence, but additionally applies $k$-means clustering in the latent space. Moreover, the method uses stacked autoencoders and does not require access to a centralized dataset. In their experimental evaluation, the authors observe strong results, but also decreased reliability in non-IID scenarios (-15\% accuracy on the MNIST dataset), despite using a robust cluster center aggregation strategy.
%F-DEC and FDEC are both designed for centralized FL (as is our method). 
In the decentralized FL setting, the work in \cite{hasan_autoencoder-based_2025} introduces an AE-based method, where clients communicate directly with each other. %The main novelty is a new aggregation method named Fednadam, evaluated on different datasets using different metrics compared to F-DEC and FDEC. 
While the authors demonstrate good performance in the IID scenario, the non-IID scenario was not part of the experimental evaluation.

In \cite{miao_contrastive_2024, miao_federated_2024}, DNN-based federated deep clustering methods were introduced. Both methods are based on contrastive loss functions and show promising results in IID data scenarios, while the experiments also uncover the impact of non-IIDness (a drop of 25\%-29\% on the clustering performance).
%\cite{miao_contrastive_2024} focuses on self-supervised federated representation learning, where clustering is used as one of the subsequent tasks to evaluate the effectiveness of the representations. 
%The authors of \cite{miao_contrastive_2024} propose a method based on contrastive learning. It achieves competitive performance in the IID scenario, but in the non-IID case, the performance drops by up to 25\% on the CIFAR-10 dataset. 
%The authors of \cite{miao_federated_2024} propose a contrastive clustering method named FedMCC. 
%During the first stage, consistent representations are learned between clients, while the second stage is focused on the clustering task. 
%The experimental evaluation shows promising results, but also uncovers the impact of non-IID data (a drop of 29\% in accuracy on the CIFAR-10 dataset).

\subsection{Handling Non-IID Data in Federated Learning}\label{sec:rel_fed_non_iid}
A key challenge in FL and federated deep clustering is data heterogeneity between clients \cite{kairouz_advances_2021, zhu_federated_2021, sana_advancing_2025} (also see previous Section \ref{sec:rel_fed_deep_clustering}). In FL, data heterogeneity, or non-IIDness, refers to the divergence of distributions between the local and private data of clients. If not mitigated, this can lead to convergence or model degradation issues.

The work in \cite{zhu_federated_2021} categorizes approaches to handle non-IIDness into three broad categories: data-, algorithm-, and system-based approaches. 
Data-based approaches address non-IIDness at the most fundamental level and aim to harmonize local data distributions through data sharing or data augmentation. While data sharing approaches, such as in \cite{zhao_federated_2018}, are highly effective, they typically assume access to a centralized (IID) dataset, which may be unavailable in privacy-preserving settings. As an alternative, it has been suggested to apply data augmentation techniques to approximate IID distributions rather than sharing raw data instances, including approaches that use synthetic data creation \cite{chen_federated_2023, cheng_gfl_2023}.

Algorithm-based approaches aim to address the non-IID problem through the learning algorithm. For example, methods such as FedProx \cite{li_federated_2020} introduce a regularization term in the local learning objective that penalizes deviation from the weights of the global model. %Other approaches like \cite{arivazhagan_federated_2019} introduce neural network architectures that consist of base and personalization layers. The authors propose to update only the base layers globally using FedAvg and optimize the personalization layers locally to allow each client to adapt the model to its local (non-IID) data.

Methods that fall into the system-based category alter the federated system. Most notably, clustered FL falls into this category. In clustered FL approaches, such as IFCA \cite{ghosh_efficient_2020}, clients with compatible data are grouped and learn one model per group instead of a single model for all clients.

Deep clustering methods solve two fundamental tasks without label supervision: representation learning and input space partitioning. 
Robust partitioning methods in non-deep clustering have been introduced that are able to deal with non-IID data. \cite{dennis_heterogeneity_2021} and \cite{stallmann_framework_2022} combine local $k$-means or fuzzy $c$-means with central server-side $k$-means averaging and demonstrate the robustness of these approaches to non-IID data.
However, learning representations in unsupervised FL remains a challenge, especially under  data heterogeneity \cite{van_berlo_towards_2020, zhang_federated_2023}.
As a solution, the authors of \cite{zhang_federated_2023} introduce dictionary and alignment modules that depend on access to a shared dataset. %While effective, access to a public dataset of raw data instances may not always be possible. 

Our approach to address non-IIDness is a hybrid of data- and algorithm-based methods.% We generate synthetic data samples to augment local data and aim to align feature spaces through the introduction of a geometry-aware regularization term in the local loss functions, as we will elaborate on in the subsequent sections.

\section{Proposed Method}\label{sec:proposed_method}
\begin{figure}
    \centering
    \includegraphics[width=0.5\linewidth]{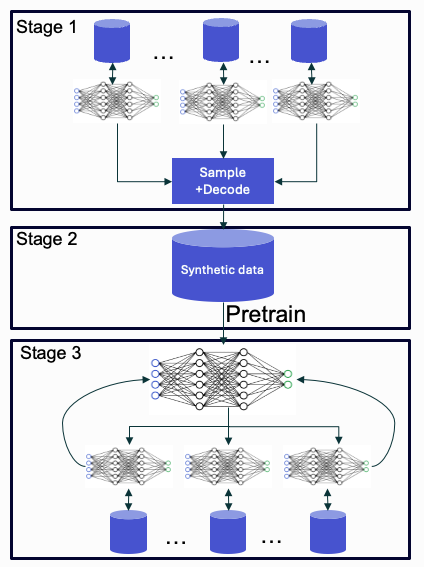}
    \caption{Illustration of the three stages in FedDCN training: synthetic data creation, pretraining, federated training.}
    \label{fig:fed_dcn_stages}
\end{figure}

The primary contribution of this work is a federated clustering method that performs well on IID and non-IID high-dimensional data, without requiring direct access to raw data instances, thereby protecting data privacy.
Previous studies have shown that AE-based federated deep clustering shows promising performance on high-dimensional IID data, but still struggles with non-IID data (see Section \ref{sec:rel_fed_deep_clustering}). To address non-IIDness during the federated training process, data-based approaches have been shown to be very effective, especially in federated unsupervised representation space learning (see Section \ref{sec:rel_fed_non_iid}).
The AE architecture of DCNs allows us to create a synthetic dataset that approximates the global data distribution without having direct access to the raw data instances. This motivates the design of our three-stage federated deep clustering framework illustrated in Figure \ref{fig:fed_dcn_stages}.

\subsection{Local Deep Clustering and Synthetic Data Creation}
%In the work of \cite{zhao_federated_2018} it has been shown that sharing even a small IID dataset with all clients stabilizes the federated training process. Similarly, the work in \cite{zhang_federated_2023} shows that a small IID dataset can be used to align feature spaces in federated unsupervised representation learning under non-IIDness.
%In our approach, an IID reference dataset is approximated with a synthetic dataset to protect the privacy of the clients' raw data instances. 
To create a synthetic IID reference dataset, we adapt the DCN model and its AE architecture.
Each client learns a DCN using its local dataset (Section \ref{sec:local_dcn}) and shares the local latent cluster centers as well as local decoders with the central server. The central server constructs a synthetic dataset by sampling around the latent space cluster centers and mapping the samples back to the observed space (Section \ref{sec:synthetic_data_creation}).

\subsubsection{Local Deep Clustering Networks - Pretraining and Alternating Optimization}\label{sec:local_dcn}
The proposed DCN consists of a stacked AE and a clustering module that operates in the latent space. Training the DCN involves optimizing the following objective:
\begin{align}
    \min_{w^{c}, [k_{l}^{c}]_{l=1}^{K}} \mathcal{L}^{c}
\end{align}
with
\begin{align}
    \mathcal{L}^{c} &= \alpha \mathcal{L}^{c}_{recon} + \lambda\mathcal{L}^{c}_{clust} + \beta \mathcal{L}^{c}_{geom}\label{eq:dcn_loss}, \\
    \mathcal{L}^{c}_{recon} &= \sum_{i=0}^{N_{c}}||x_{i}-g(f(x_{i}))||_{2}^{2}, \\
    \mathcal{L}^{c}_{clust} &= \sum_{l=1}^{K}\sum_{i=0}^{N_{c}}\mathds{1}_{x_{i}\text{ assigned to }{k_{l}}} ||f(x_{i}) - k_{l}||_{2}^{2} \\
    \mathcal{L}^{c}_{geom}&= \sum_{i=0}^{N_{c}} ||f(x_{i})-umap(x_{i})||_{2}^{2}\label{eq:loss_geom},
\end{align}
where $w^{c}$ are the weights of client $c$'s AE network, $[k^{c}]_{l=1}^{K}$ are client $c$'s latent space cluster centers, $K$ is the number of clusters, and $N_{c}=|X^{c}|$ is the number of data points of client $c$. 
The AE's encoder layers are stacked on top of each other, a bottleneck layer is added, and the encoder layers are mirrored  to define the decoder layers. Each layer applies batch normalization and uses ReLu as the activation function.
Note that we introduce a geometric loss term $\mathcal{L}^{c}_{geom}$ that the original DCN formulation does not contain. This is to prevent distortions in the latent space (Section \ref{sec:geom_regularization}).

The optimization algorithm is similar to the one introduced by \cite{yang_towards_2017}. First, the network is pretrained with $\lambda=0$ (that is, ignoring the clustering loss) for $T_{pre}$ epochs using the Adam optimizer \cite{kingma_adam_2014}. Second, an alternating optimization algorithm minimizes $\mathcal{L}^{c}$. The algorithm alternates between two steps: updating the network parameters $w^{c}$ with fixed cluster centers $[k^{c}]_{l=1}^{K}$ and updating cluster centers and assignments with fixed $w^{c}$ (see Algorithm \ref{alg:local_dc} for an overview).

\begin{algorithm}[]
\caption{Local DCN optimization}
\label{alg:local_dc}
\begin{algorithmic}[1]
\Statex
    // Pretraining
    \For{$t=1, \dots, T_{pre}$}                    
        \State {$w^{c}_{t}$ $\gets$ {$Adam(w^{c}_{t-1})$}} // minimizing (\ref{eq:dcn_loss}) with $\lambda=0$
    \EndFor \\
    // DCN training
    \For{$t=1, \dots, T$}                    
        \State {$w^{c}_{t}$ $\gets$ {$Adam(w^{c}_{t-1})$}} // minimizing (\ref{eq:dcn_loss}) with $\lambda\ge0$
        \State {$[k^{c}]_{l=1}^{K}$ $\gets$ $kmeans(f_{w_{t}^{c}}(X^{c}))$}
        \State{Update cluster assignments}
    \EndFor
\end{algorithmic}
\end{algorithm}

After local DCN optimization, each client computes the standard deviations of the distances between each cluster center and the points assigned to it. Every client $c=1,\dots, C$ shares standard deviations $\sigma_{l}^{c}$ of all clusters $l=1,\dots, K$ with the central server, where these standard deviations inform the latent space sampling (Section \ref{sec:synthetic_data_creation}).

\subsubsection{Synthetic Data Creation - Sample and Decode}\label{sec:synthetic_data_creation}
Given all clients' latent space cluster centers, decoders, and per-cluster standard deviations, the central server constructs the synthetic dataset to be used for pretraining and latent space alignment.
Given a client $c$, a cluster center $k_{l}^{c}$, and the corresponding standard deviation $\sigma_{l}^{c}$, the central server draws $N_{l}^{synth}$ latent space samples from an isotropic Gaussian centered at $k_{l}$ with covariance $\Sigma_{l}^{c}=\sigma_{l}^{c}*I$, where $I$ is the identity matrix. Then, the synthetic dataset is created by decoding the latent samples using client $c$'s decoder $g_{c}$:
\begin{align}\label{eq:synthetic_data_creation}
    X^{synth} &= \bigcup_{c=1}^{C}\bigcup_{l=1}^{K} \{g_{c}(x_{latent}): x_{latent}\sim\mathcal{N}(k_{l}^{c}, \Sigma_{l}^{c}) \}.
\end{align}

Note that assuming isotropic Gaussians is a simplification that likely introduces inaccuracies in the latent space sampling, but also protects clients' data privacy as it reveals relatively little information about clients' latent space distributions.

In our experiments, we fix the synthetic dataset size to 5000 samples. Each client's decoder is used to generate a number of samples proportional to its local dataset size. The number of samples per cluster center $N_{l}^{synth}$ is derived from the number of samples within a radius of one standard deviation of the cluster center and is proportional to it. That way, we prioritize each of the clients' most cohesive centers and deprioritize less cohesive centers, indicating a weaker data partition.
\subsubsection{Geometry Regularization}\label{sec:geom_regularization}
Injecting geometric information into the loss function serves two main purposes. Firstly, it can prevent latent space distortions \cite{nazari_geometric_2023}. Since we sample in the latent space within a radius of one standard deviation around cluster centers, these distortions may lead to imbalanced or non-IID synthetic data (thereby defeating their purpose of providing alignment in the non-IID scenario). 
Secondly, the geometric loss introduces latent feature space alignment between clients during the federated training process (Section \ref{sec:fed_training}). The central server aligns the local representation spaces by injecting the same geometric information into every client's local learning objective. It shares synthetic datapoints with their geometric embeddings, and the clients include the geometry regularization (Equation (\ref{eq:loss_geom})) in the local learning objectives. This is similar to aligning feature spaces as in \cite{zhang_federated_2023}, without accessing raw data instances.

To inject geometric information, we follow the approach of \cite{duque_geometry_2023} and introduce a regularization term $\mathcal{L}^{c}_{geom}$ (Equation (\ref{eq:loss_geom})) in the loss function. For each feature vector $x$, its geometric embedding $umap(x)$ is computed, and large distances between the learned and geometric embeddings are penalized.
UMAP (Uniform Manifold Approximation and Projection) is a manifold learning technique that maps data from a high-dimensional space to a low-dimensional space while preserving the (assumed) geometric structure and pairwise distances \cite{mcinnes_umap_2018}. Thus, $umap:\mathbb{R}^{p}\to \mathbb{R}^{d}$ denotes the function that maps a high-dimensional feature vector $x\in X^{c}$ to its geometric $d$-dimensional embedding. 

\subsection{Initialization and Pre-Training}
The central server initializes the global model by pretraining it on the synthetic dataset $X^{synth}$ and its geometric embeddings $X^{geom}= \{umap(x) : x\in X^{synth}\}$.

After creating $X^{geom}$, Algorithm \ref{alg:local_dc} is applied to minimize Equation (\ref{eq:dcn_loss}). 
At the end of this phase, the central server shares global model weights $w_{0}$, cluster centers $[k_{l}]_{l=1}^{K}$, and datasets $X^{synth}$ and $X^{geom}$ with the clients.

\subsection{Federated Simultaneous Optimization of Representations and Clustering}\label{sec:fed_training}

In the final stage, the clients collaboratively train the DCN and minimize: $\min_{w, [k_{l}]_{l=1}^{K}} \sum_{c=1}^{C}\frac{1}{N_{c}}\mathcal{L}^{c}$.
Similarly to local DCN optimization, federated optimization alternates between updating global autoencoder weights $w_{t}$ with fixed cluster centers $[k_{l}]_{l=1}^{K}$ and updating cluster centers with fixed autoencoder weights for a predefined number of epochs $T_{fed}$.

In the AE weights update phase in epoch $t$, each client $c$ first optimizes Equation (\ref{eq:dcn_loss}) locally using the Adam optimizer. Note that each batch of the clients' local dataset is augmented with a batch of the same size from the shared synthetic dataset $X^{synth}$. After the local optimization, the resulting local AE weights $w_{t}^{c}$ are shared with the central server, which applies federated averaging to derive the new global AE weights $w_{t} \gets \sum_{c=1}^{C}\frac{N_{c}}{N}w_{t}^{c}$, where $N=\sum_{c=1}^{C}N_{c}$ is the total number of datapoints. Then, the central server shares the updated $w_{t}$.

Using the updated global weights, each client $c$ maps its data into the new latent space, where it applies $kmeans$ to identify new local cluster centers $[k_{l}^{c}]_{l=1}^{K}$. After collecting all local latent cluster centers, the central server runs $kmeans$ on them to derive the new global latent space cluster centers $[k_{l}]_{l=1}^{K}$ and shares them with the clients.
The federated training process is outlined in Algorithm \ref{alg:fed_dcn}.

\begin{algorithm}[]
\caption{FedDCN training protocol}
\label{alg:fed_dcn}
\begin{algorithmic}[1]
\Statex
    \State{Local pretraining according to Algorithm \ref{alg:local_dc}}.
    \State{Create $X^{synth}$ according to Equation (\ref{eq:synthetic_data_creation}).}
    \State{Create $X^{geom}$}
    \State{ $w_{0}, [k_{l}]_{l=1}^{K}$ $\gets$ Algorithm \ref{alg:local_dc} with $X^{synth}$ // Initialize global model AE weights and cluster centers on the central server}
    \State{broadcast($w_{0}, [k_{l}]_{l=1}^{K}, X^{synth}, X^{geom}$)}
    \For{$t=1, \dots, T_{fed}$}
    \State{$weights \gets \{\}$}
    \For{$client\in\{1,\dots, C\}$}
    \State{$w^{c}_{t} \gets client.LocalAeUpdate()$}
    \State{$weights \gets weights\cup \{ w^{c}_{t} \}$}
    \EndFor
    \State{$w_{t}\gets \sum_{c=1}^{C}\frac{N_{c}}{N}w_{t}^{c}$}
    \State{broadcast($w_{t}$)}
    \State{$localCenters \gets \{\}$}
    \For{$client\in \{1, \dots, C\}$}
    \State{$clientCenters$ $\gets$ $client.LocalClustering()$}
    \State{$localCenters$ $\gets$ $localCenters\cup \{ clientCenters \}$}
    \EndFor
    \State{$[k_{l}]_{l=1}^{K} \gets kmeans(localCenters)$}
    \State{broadcast($[k_{l}]_{l=1}^{K}$)}
    \EndFor
\end{algorithmic}
\end{algorithm}
\section{Experimental Evaluation}\label{sec:exp_eval}
This section evaluates our proposed method with respect to the following criteria: effectiveness in comparison to related federated deep clustering methods, impact of non-IID data, impact of an increasing number of clients (or, decreasing the number of local datapoints as the dataset is fixed), and sensitivity to initialization and hyperparameter choice.

The following subsections describe the experimental design before describing and discussing the results. Code to replicate the experiments is available on GitHub: \url{https://github.com/stallmo/fed-dcn}.

\subsection{Data}
In the centralized setting, DCNs were developed as an enhancement of DECs and, hence, our primary goal is to compare FedDCN with the federated versions of DECs, namely F-DEC and FDEC (see Section \ref{sec:rel_fed_deep_clustering}). %FDEC is evaluated on the standard benchmark datasets MNIST, Fashion-MNIST, USPS, and Reuters datasets as well as a breast ultrasound image dataset while F-DEC is evaluated on the MNIST and USPS datasets as well as a dataset of GPS trajectories.

In our experiments, the proposed method is trained on the same image datasets as F-DEC and FDEC: MNIST, Fashion-MNIST, and USPS, and evaluated on the standard test splits to be able to compare our results with F-DEC and FDEC.

To create federated IID and non-IID scenarios, we follow the standard protocol of \cite{li_federated_2022, zhu_federated_2021}. Specifically, the training data are distributed across clients according to the Dirichlet distribution, based on class labels with concentration parameters $\alpha_{dirichlet}=1000.0$ (approximate IID) and $\alpha_{dirichlet}=0.5$ (non-IID). 
All metrics are calculated on the unseen test dataset.

\subsection{Evaluation Metrics}
The related deep clustering methods F-DEC and FDEC are evaluated using clustering accuracy (ACC) and normalized mutual information (NMI), which we also include in our evaluation:
\begin{align}
    ACC &= \frac{\max_{\pi} \sum_{k} confusion[k, \pi(k)]}{N}, \label{eq:acc}\\
    NMI(Y,C) &= \frac{I(Y, C)}{\frac{1}{2}[H(Y)+H(C)]}. \label{eq:nmi}
\end{align}

where $confusion$ is the confusion matrix of the assignment of clusters and the ground truth labels and $\pi$ is a one-to-one mapping between the assignment of the cluster and the labels, $N$ is the total number of points in the test set.
NMI measures the agreement between the set of ground truth assignments $Y$ and the cluster assignments $C$ and normalizes it by the mean entropy of the sets. $I(\cdot, \cdot)$ denotes the mutual information score and $H(\cdot)$ is the entropy \cite{murphy_probabilistic_2022}.

Since the calculation of these external cluster validation metrics relies on the availability of ground truth labels (which may not be available in real-world clustering), we also include an internal cluster validation metric in our evaluation. Unlike validation metrics that require the calculation of pairwise distances, the Davies-Bouldin index (DBI) can easily be calculated in a federated setting \cite{stallmann_framework_2022}: $DBI = \frac{1}{K}\sum_{i=1}^{K}R_{i}$, 
where $R_{i} := \max_{i\ne j}R_{ij}$, $R_{ij} = \frac{S_{i}+S_{j}}{M_{ij}}$, $S_{i} = \frac{1}{N}\sum_{j=1}^{N}||x_{j}-c_{i}||$ is the ``cluster spread", and
$M_{ij} = (\sum_{k=1}^{D}||c_{i}[k]-c_{j}[k]||)$ is the ``center separation". Intuitively, good clusters $i$ are cohesive (low spread $S_{i}$) and well separated (high $M_{ij}\forall j$). Hence, the lower the DBI, the better the clustering.

Note that our results are optimized for accuracy to facilitate comparison with existing works. In practice, ground truth label information may not be available, and optimization for internal validation metrics like the DBI may be more appropriate.
In additional experiments, we observe that the DBI can be improved at the expense of lower $ACC$ and $NMI$.

\subsection{Experimental Protocol}
The experimental evaluation follows three steps per data scenario (dataset and its distribution across clients). First, we search for the hyperparameters that give the best performance in terms of accuracy according to Equation (\ref{eq:acc}) on the train set, then calculate all metrics in the five-client scenario, and lastly calculate the 20-client scenario metrics.
First, given a data scenario, we apply hyperparameter optimization (HO) for 50 rounds. The HO is implemented using Optuna \cite{akiba_optuna_2019}. After this step, the optimal hyperparameters are fixed.
Second, with the optimal hyperparameters, we repeat the same experiment five times with different seeds due to the sensitivity of the clustering algorithms to initialization. We report both the best and the average results. All reported metrics are calculated on the test set.
Lastly, the number of clients is increased from five to 20, as in the experimental evaluation of F-DEC and FDEC. Since the dataset and its size are fixed, this leads to every client having fewer data to learn from, potentially leading to weaker AEs for synthetic data creation and a negative impact on clustering performance. The best and average results are reported.

\begin{table}[]
\centering
\caption{Hyperparameters search space.}
\resizebox{0.7\columnwidth}{!}{%
\begin{tabular}{l|l}
Hyperparameter & Search Space \\ \hline
AE layer dimensions & \begin{tabular}[c]{@{}l@{}} \{ (128), (256, 128), \\ (512, 256), (512, 256, 128), \\ (1024, 512, 256, 128, 64) \}\end{tabular} \\
Bottleneck dimension $d$ & $[3, 16]$ \\
$\beta^{pre}$ during pretraining & $[0.0001, 1.0]$ \\
$\beta^{fed}$ during federated training & $[0.0001, 1.0]$ \\
$\lambda$ & $[0.0001, 1.0]$ \\
$\alpha$ & $[0.1, 10.0]$
\end{tabular}%
}
\label{tab:hyperparameters}
\end{table}

\begin{table}[]
\centering
\caption{Best results on the MNIST, Fashion-MNIST, and USPS datasets (avg. over five runs using the same hyperparameters in parantheses). *Results are obtained using a different latent space sampling strategy during synthetic data creation due to the small dataset size.}
\resizebox{\columnwidth}{!}{%
\begin{tabular}{lllll}
\hline
\multicolumn{5}{l}{MNIST} \\ \hline
Clients & \multicolumn{1}{l|}{Scenario} & Accuracy & NMI & DBI (latent space) \\ \hline
\multirow{2}{*}{5} & \multicolumn{1}{l|}{IID} & \begin{tabular}[c]{@{}l@{}}0.8026 \\ (0.7643 $\pm 0.0525$)\end{tabular} & \begin{tabular}[c]{@{}l@{}}0.7592 \\ (0.7445 $\pm 0.0144$)\end{tabular} & \begin{tabular}[c]{@{}l@{}}0.6915 \\ (0.7395 $\pm 0.0391$)\end{tabular} \\
 & \multicolumn{1}{l|}{Non-IID} & \begin{tabular}[c]{@{}l@{}}0.7799 \\ (0.7328 $\pm 0.0515$)\end{tabular} & \begin{tabular}[c]{@{}l@{}}0.7922 \\ (0.7693 $\pm 0.0221$)\end{tabular} & \begin{tabular}[c]{@{}l@{}}0.9905 \\ (1.0758 $\pm 0.0826$)\end{tabular} \\ \hline
\multirow{2}{*}{20} & \multicolumn{1}{l|}{IID} & \begin{tabular}[c]{@{}l@{}}0.7160\\ (0.6838 $\pm 0.0206$)\end{tabular} & \begin{tabular}[c]{@{}l@{}}0.6772 \\ (0.6666 $\pm 0.0103$)\end{tabular} & \begin{tabular}[c]{@{}l@{}}0.7462 \\ (0.7929 $\pm 0.0396$)\end{tabular} \\
 & \multicolumn{1}{l|}{Non-IID} & \begin{tabular}[c]{@{}l@{}}0.7636 \\ (0.7029 $\pm 0.0475$)\end{tabular} & \begin{tabular}[c]{@{}l@{}}0.7442 \\ (0.6981 $\pm 0.0423$)\end{tabular} & \begin{tabular}[c]{@{}l@{}}1.066 \\ (1.1555 $\pm 0.0925$)\end{tabular} \\ \hline
\multicolumn{5}{l}{Fashion-MNIST} \\ \hline
Clients & \multicolumn{1}{l|}{Scenario} & Accuracy & NMI & DBI (latent space) \\ \hline
\multirow{2}{*}{5} & \multicolumn{1}{l|}{IID} & \begin{tabular}[c]{@{}l@{}}0.6427 \\ (0.5866 $\pm 0.0443$)\end{tabular} & \begin{tabular}[c]{@{}l@{}}0.6491 \\ (0.6164 $\pm 0.0183$)\end{tabular} & \begin{tabular}[c]{@{}l@{}}0.1422\\ (0.3504$\pm 0.1597$)\end{tabular} \\
 & \multicolumn{1}{l|}{Non-IID} & \begin{tabular}[c]{@{}l@{}}0.5885 \\ (0.5650 $\pm 0.0253$)\end{tabular} & \begin{tabular}[c]{@{}l@{}}0.6103 \\ (0.6022 $\pm 0.0090$)\end{tabular} & \begin{tabular}[c]{@{}l@{}}0.6816 \\ (0.8389 $\pm 0.1238$)\end{tabular} \\ \hline
\multirow{2}{*}{20} & \multicolumn{1}{l|}{IID} & \begin{tabular}[c]{@{}l@{}}0.6028 \\ (0.5849 $\pm 0.0117$)\end{tabular} & \begin{tabular}[c]{@{}l@{}}0.6193 \\ (0.6003 $\pm 0.0182$)\end{tabular} & \begin{tabular}[c]{@{}l@{}}0.3049 \\ (0.4184 $\pm 0.116$)\end{tabular} \\
 & \multicolumn{1}{l|}{Non-IID} & \begin{tabular}[c]{@{}l@{}}0.5561 \\ (0.5241 $\pm 0.0318$)\end{tabular} & \begin{tabular}[c]{@{}l@{}}0.5956 \\ (0.5720 $\pm 0.0269$)\end{tabular} & \begin{tabular}[c]{@{}l@{}}0.8395 \\ (0.8853 $\pm 0.0452$)\end{tabular} \\ \hline
\multicolumn{5}{l}{USPS} \\ \hline
Clients & Scenario & Accuracy & NMI & DBI (latent space) \\ \hline
\multirow{2}{*}{5} & \multicolumn{1}{l|}{IID} & \begin{tabular}[c]{@{}l@{}}0.7848\\ (0.7521 $\pm 0.0448$)\end{tabular} & \begin{tabular}[c]{@{}l@{}}0.7472\\ (0.7323 $\pm 0.0164$)\end{tabular} & \begin{tabular}[c]{@{}l@{}}0.3812\\ (0.576 $\pm 0.1546$)\end{tabular} \\
 & \multicolumn{1}{l|}{Non-IID} & \begin{tabular}[c]{@{}l@{}}0.6956\\ (0.6239 $\pm 0.0577$)\end{tabular} & \begin{tabular}[c]{@{}l@{}}0.7364\\ (0.7100 $\pm 0.0280$)\end{tabular} & \begin{tabular}[c]{@{}l@{}}0.5257\\ (0.6256 $\pm 0.1066$)\end{tabular} \\ \hline
\multirow{2}{*}{20} & \multicolumn{1}{l|}{IID} & \begin{tabular}[c]{@{}l@{}}0.7653*\\ (0.6994 $\pm 0.0597$)\end{tabular} & \begin{tabular}[c]{@{}l@{}}0.7036*\\ (0.6825 $\pm 0.0095$)\end{tabular} & \begin{tabular}[c]{@{}l@{}}0.3177*\\ (0.3930 $\pm 0.0579$)\end{tabular} \\
 & \multicolumn{1}{l|}{Non-IID} & \begin{tabular}[c]{@{}l@{}}0.7803*\\ (0.7159 $\pm 0.0372$)\end{tabular} & \begin{tabular}[c]{@{}l@{}}0.7076*\\ (0.6925 $\pm 0.0124$)\end{tabular} & \begin{tabular}[c]{@{}l@{}}0.8780*\\ (0.9764$\pm 0.0634$)\end{tabular}
\end{tabular}%
}
\label{tab:datasets_results}
\end{table}

\subsection{Results}\label{sec:results}
This section presents and discusses the experimental results, organized by the main research questions. All results of our experiments can be found in Table \ref{tab:datasets_results}, a comparison with F-DEC and FDEC can be found in Table \ref{tab:result_comparison}, and the optimal hyperparameters in terms of accuracy in Table \ref{tab:opt_hyper}.

\subsubsection{Comparison with related deep clustering methods}
The proposed method is compared with F-DEC and FDEC. We report the best results of our method and compare them to the results as reported in the original evaluations of F-DEC and FDEC, as can be seen in Table \ref{tab:result_comparison}.

We observe similar performance in most scenarios. In particular, in the IID scenarios, FedDCN's performance matches the reported performance of both F-DEC and FDEC in most cases.
In the non-IID scenarios, a notable performance difference can be observed in the MNIST experiments. While F-DEC and FDEC are negatively influenced by data heterogeneity, FedDCN achieves even better performance than in the IID scenario.
In the Fashion-MNIST experiments, the accuracy and NMI of FedDCN drop slightly more than those of FDEC. In absolute terms, the performance is still almost identical.
On the USPS dataset, both FDEC and FedDCN are not strongly affected by data heterogeneity. However, FDEC achieves a better NMI score in both scenarios. F-DEC is outperformed by both FDEC and FedDCN by a larger margin.

In summary, the methods have different strengths and weaknesses. 
On the one hand, the proposed method FedDCN is the only one that demonstrates strong non-IID robustness across all three benchmark datasets.
This can be explained by the different approaches to handle data heterogeneity. While FDEC relies on robust cluster center aggregation, FedDCN also applies synthetic data augmentation and a geometry-aware loss term, leading to improved robustness.
On the other hand, FedDCN achieves worse NMI on some datasets like USPS. This can be attributed to the geometry loss term.
In particular, the optimal geometric loss weight during pretraining is the highest in the USPS experiments (see Table \ref{tab:opt_hyper}), potentially enforcing $umap$ embedding alignment too strongly. Moreover, the hyperparameter search optimized accuracy and not NMI, possibly leading to suboptimal NMI performance.

\begin{table}[]
\centering
\caption{Performance comparison of FedDCN (ours) and the results as reported by F-DEC \cite{mashhadi_deep_2021} and FDEC \cite{xu_federated_2025} in federated IID and non-IID scenarios with 20 clients.}
\resizebox{0.55\columnwidth}{!}{%
\begin{tabular}{lllll}
\cline{3-5}
 &  & F-DEC & FDEC & FedDCN \\ \hline
\multicolumn{5}{l}{MNIST IID} \\ \hline
 & ACC & - & \textbf{0.74} & 0.72 \\
 & NMI & 0.65 & \textbf{0.68} & \textbf{0.68} \\ \hline
\multicolumn{5}{l}{MNIST Non-IID} \\ \hline
 & ACC & 0.57 & 0.63 & \textbf{0.76} \\
 & NMI & 0.48 & 0.58 & \textbf{0.74} \\ \hline
\multicolumn{5}{l}{USPS IID} \\ \hline
 & ACC & - & \textbf{0.79} & 0.77 \\
 & NMI & - & \textbf{0.82} & 0.70 \\ \hline
\multicolumn{5}{l}{USPS Non-IID} \\ \hline
 & ACC & 0.68 & \textbf{0.78} & \textbf{0.78} \\
 & NMI & 0.54 & \textbf{0.79} & 0.71 \\ \hline
\multicolumn{5}{l}{Fashion-MNIST IID} \\ \hline
 & ACC & - & 0.59 & \textbf{0.60} \\
 & NMI & - & 0.61 & \textbf{0.62} \\ \hline
\multicolumn{5}{l}{Fashion-MNIST Non-IID} \\ \hline
 & ACC & - & \textbf{0.58} & 0.56 \\
 & NMI & - & \textbf{0.60} & \textbf{0.60}
\end{tabular}%
}
\label{tab:result_comparison}
\end{table}

\subsubsection{Impact of Non-IID Data}
In the MNIST dataset experiments with five clients, the best accuracy drops from $0.80$ to $0.78$. On the contrary, the NMI increases with the introduction of non-IIDness. With 20 clients, accuracy and NMI both increase in the non-IID scenario.
On the Fashion-MNIST dataset, a negative impact of non-IIDness can be observed in experiments with five and 20 clients in both metrics. With five clients, the accuracy drops from $0.64$ to $0.59$ and with 20 clients from $0.60$ to $0.56$ while the NMI is less impacted.
Lastly, the experiments with the USPS dataset show an impact on accuracy only in experiments with five clients. The NMI is not affected in experiments with five or 20 clients. Furthermore, the accuracy even increases in the non-IID experiments with 20 clients. 
Notably, the DBI increases (indicating poorer data partitions) in all cases where data heterogeneity is introduced. We attribute that to the fact that we optimize for an unrelated metric and expect the effect to be weaker when optimizing for DBI.

Overall, the impact of non-IIDness is small, and sometimes even positive, which suggests that the combination of synthetic data augmentation and geometry-aware regularization can help addressing non-IID issues in federated deep clustering.
%In general, there are two levers to further improve robustness: increasing the weight of the geometry loss term or improving the synthetic dataset. Given the rather low optimal weight of the geometric loss term $\beta$ in most experiments (Table \ref{tab:opt_hyper}), improving the synthetic data creation is a more promising direction to further improve robustness, e.g., through better latent space sampling strategies or a dedicated synthetic data creation model (rather than repurposing the DCN architecture).

\subsubsection{Increasing Number of Clients}
In our experiments, we observe a drop in clustering performance as the number of clients increases, similar to the observations in \cite{mashhadi_deep_2021} and \cite{xu_federated_2025}.
On the one hand, the accuracy decreases sharply from $0.8$ to $0.72$ in the MNIST IID scenario when increasing the number of clients from five to 20.
On the other hand, accuracy decreases only from $0.78$ to $0.76$ in the MNIST non-IID experiments. 
In the Fashion-MNIST scenarios, the impact on accuracy is between $-5\%$ and $-7\%$ relative performance. Similarly, the impact on NMI is between $-12\%$ (MNIST IID) and $-3\%$ (Fashion-MNIST non-IID). 
In the USPS experiments, only a light impact is observed. The accuracy and DBI improve in the IID and regresses in the non-IID scenarios while the NMI slightly decreases in both.

Note that we had to adjust the data sampling strategy in the synthetic data creation step after stage 1 in the USPS experiments with 20 clients. The low number of data points per client and latent space sampling strategy (Section \ref{sec:synthetic_data_creation}) led to multiple zero cluster counts and a misaligned synthetic dataset. Rather than deciding the number of points to sample per cluster center based on the number of points within a radius of one standard deviation around the centers, the number of points is set to the total number of points assigned to the cluster center. We also observe a positive effect of that new sampling strategy on non-IIDness performance with five clients in the USPS experiments but leave a more principled investigation for future work.
%This adjustment hints at a potential mitigation strategy against the negative impact of increased number of clients or non-IIDness: improving the synthetic data creation leads to smaller impact of increasing the number of clients (or, put differently: decreasing the number of points per client). 

\subsubsection{Sensitivity to Initialization}
Clustering algorithms, such as $k$-means are sensitive to cluster center initialization and the randomness it introduces \cite{harris_extensive_2022}. FedDCN initializes $k$-means multiple times, both locally in stage 1 and globally in stage 2. Moreover, neural networks are also known to be susceptible to initialization.
To study the influence of initialization in FedDCN, each optimal run is repeated five times with the same hyperparameters but different random seeds.

Across all experiments, we observe a noticeable, but moderate effect of randomness. For example, in the USPS experiments with 20 clients, the best run ($0.7653$ accuracy) is significantly better than the average over five runs ($0.6994$). In other cases, such as the IID Fashion-MNIST experiments with 20 clients, the best performance ($0.6028$) is close to the average over the five runs ($0.5849$). The variability in NMI is consistently lower than the variability in accuracy while DBI varies the most, presumingly because it is completely unrelated to the hyperparameter optimization objective.
In summary, it is advisable to repeat the training multiple times due to randomness during initialization and training. 

\subsubsection{Impact of Hyperparameters}
\begin{figure}
    \centering
    \includegraphics[width=0.87\linewidth]{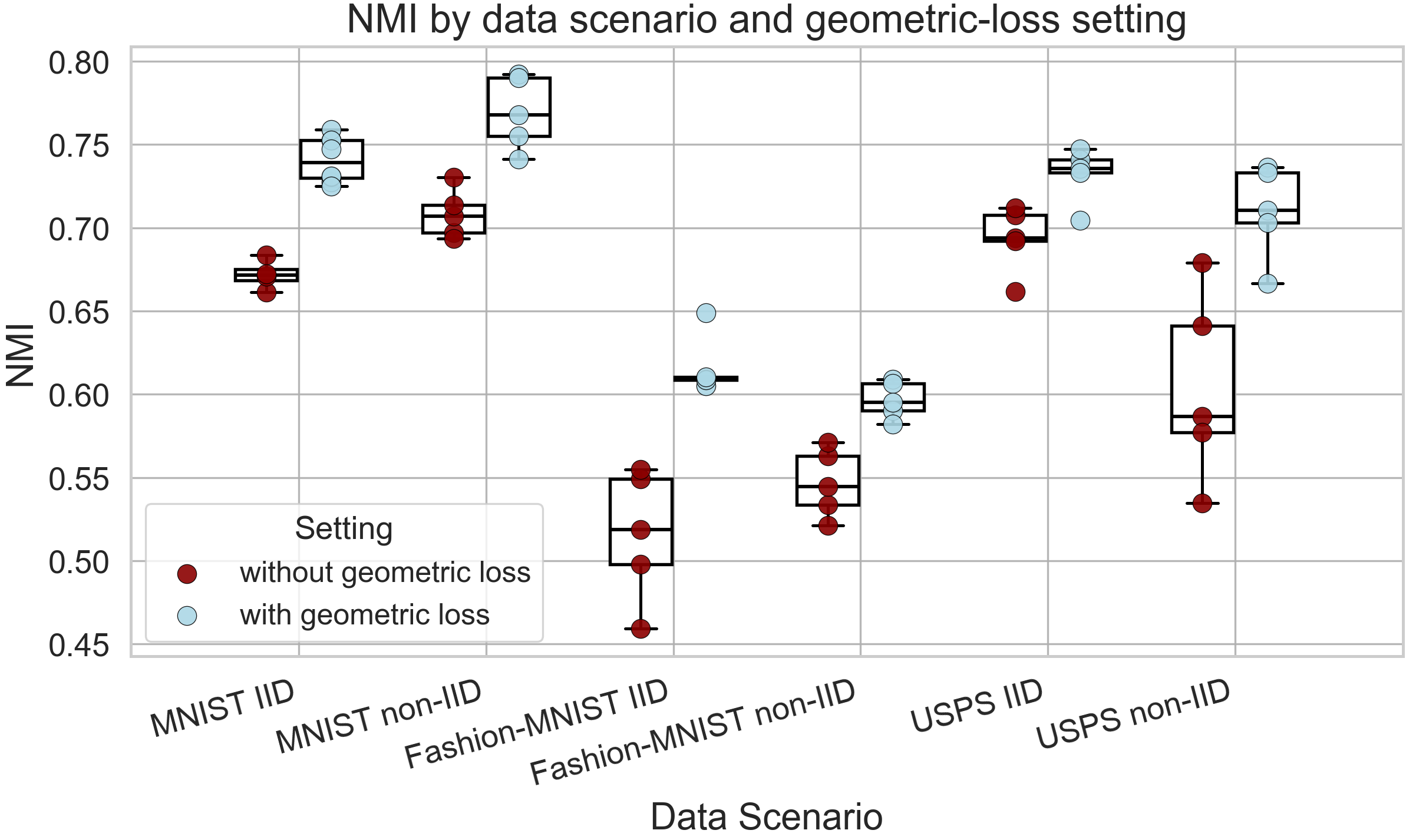}
    \caption{Effect of removing the geometric loss on the NMI.}
    \label{fig:nmi_ablation}
\end{figure}

As Table \ref{tab:hyperparameters} shows, optimal hyperparameters vary widely between different data scenarios, highlighting the necessity of hyperparameter optimization. 
For example, the best geometric loss weight $\beta$ during pretraining on the USPS dataset is $0.3711$ in the IID scenario and $0.1970$ in the non-IID scenario while it is $\le0.0304$ in the remaining experiments. 
Similarly, the reconstruction error weight $\alpha$ and cluster loss weight $\lambda$ vary by multiple orders of magnitude.

Furthermore, an ablation study is conducted to understand the importance of including the geometric regularization term in the loss function. The experiments with five clients are repeated with setting $\beta^{pre}=\beta^{fed}=0$ and keeping the remaining best hyperparameters fixed for every data scenario. Including the geometric loss has a positive impact on all metrics in all data scenarios with the exception of the DBI in the USPS IID and Fashion-MNIST IID scenarios. Whether this effect persists when optimizing for the DBI is an interesting follow-up research question. The positive effect on NMI (which is illustrated in Figure \ref{fig:nmi_ablation}) is slightly stronger than on ACC. The ACC and NMI performance gains in all data scenarios let us conclude that the geometric loss is an important component of the method to align local feature spaces and avoid latent space distortions.
However, further ablation studies will enhance understanding of the components' interplay and the contribution of each loss term in different IID and non-IID scenarios.
%In contrast, the geometric loss weight $\beta$ during federated training is relatively small in most scenarios. Especially in the non-IID scenarios, it is $\le 0.0017$, which may indicate that the synthetic data augmentation is more important to ensure robustness against data heterogeneity.

\begin{table}[]
\centering
\caption{Optimal hyperparameters (optimized for accuracy).}
\resizebox{\columnwidth}{!}{%
\begin{tabular}{l|lllllll}
Data Scenario & AE layers & \begin{tabular}[c]{@{}l@{}} $d$ \end{tabular} & \begin{tabular}[c]{@{}l@{}}$\beta^{pre}$\end{tabular} & \begin{tabular}[c]{@{}l@{}}$\beta^{fed}$\end{tabular} & $\lambda$ & $\alpha$ & \begin{tabular}[c]{@{}l@{}}Learn. \\ rate\end{tabular} \\ \hline
MNIST IID & {[}256, 128{]} & 3 & 0.0037 & 0.0204 & 0.0002 & 0.8273 & 0.0086 \\
MNIST non-IID & {[}512, 256, 128{]} & 4 & 0.0076 & 0.0003 & 0.0003 & 2.8550 & 0.0014 \\
Fashion IID & {[}512, 256, 128{]} & 6 & 0.0131 & 0.0003 & 0.2814 & 0.2137 & 0.0008 \\
Fashion non-IID & {[}512, 256{]} & 6 & 0.0304 & 0.0008 & 0.0005 & 0.3225 & 0.0001 \\
USPS IID & \begin{tabular}[c]{@{}l@{}}{[}1024, 512, \\ 256, 128, 64{]}\end{tabular} & 4 & 0.3711 & 0.0241 & 0.0470 & 0.1523 & 0.0002 \\
USPS non-IID & {[}512, 256{]} & 5 & 0.1970 & 0.0017 & 0.0006 & 0.5828 & 0.0024
\end{tabular}%
}
\label{tab:opt_hyper}
\end{table}

\section{Concluding Remarks}\label{sec:concluding_remarks}
This work introduces FedDCN, a novel federated deep clustering method, that generalizes the concept of Deep Clustering Networks to the FL setting. The proposed method achieves comparable performance to other federated deep clustering techniques in the IID setting and favorable robustness against data heterogeneity. That robustness stems from a synthetic data augmentation protocol and the introduction of a geometry-regularized loss term. 
Our experimental evaluation reveals sensitivity to hyperparameter choices and hints to improvements through different latent space sampling strategies. An ablation study shows the importance of including the geometric loss component and motivates an extended ablation study to better understand the impact of further components in FedDCN. Moreover, future work will focus on evaluation on additional datasets and explore the application to drift or anomaly detection, possibly using alternative AE architectures.

\section*{Acknowledgment}
The authors disclose the use of Claude Code to assist in the implementation of the experiments. They take full responsibility for the experimental evaluation and the content of the paper.

\bibliographystyle{IEEEtran}

\bibliography{references/references_unlinked_feddcn}

\end{document}